\pdfoutput=1

\documentclass[11pt]{article}

\usepackage{acl}

\usepackage{times}
\usepackage{latexsym}
\usepackage{amsmath}
\usepackage[T1]{fontenc}

\usepackage[utf8]{inputenc}
\usepackage[inline]{enumitem}
\usepackage{booktabs}
\usepackage{graphicx}

\usepackage{microtype}

\usepackage{hyperref}

\newcommand{\qid}[1]{\href{https://www.wikidata.org/wiki/#1}{#1}}
\newcommand{\todo}[1]{{}}
\title{Robust Candidate Generation for Entity Linking on Short Social Media Texts}

\author{Liam Hebert \\
  University of Waterloo \\
  \texttt{l2hebert@uwaterloo.ca} \\\And
  Raheleh Makki \\
  Twitter \\
  \texttt{rmakki@twitter.com} \\ \And
  Shubhanshu Mishra \\
  Twitter \\
  \texttt{smishra@twitter.com} \\ \AND
  Hamidreza Saghir \\
  Twitter \\
  \texttt{hsaghir@twitter.com} \\ \And 
  Anusha Kamath \\
  Twitter \\
  \texttt{akamath@twitter.com} \\ \And 
  Yuval Merhav \\
  Twitter \\
  \texttt{ymerhav@twitter.com}
  }
\begin{document}
\maketitle
\begin{abstract}

Entity Linking (EL) is the gateway into Knowledge Bases. Recent advances in EL utilize dense retrieval approaches for Candidate Generation, which addresses some of the shortcomings of the Lookup based approach of matching NER mentions against pre-computed dictionaries. In this work, we show that in the domain of Tweets, such methods suffer as users often include informal spelling, limited context, and lack of specificity, among other issues. We investigate these challenges on a large and recent Tweets benchmark for EL, empirically evaluate lookup and dense retrieval approaches, and demonstrate a hybrid solution using long contextual representation from Wikipedia is necessary to achieve considerable gains over previous work, achieving 0.93 recall.
\end{abstract}

\section{Introduction}
Entity Linking (EL) is the task of linking mentions to their corresponding entities in a Knowledge Base (KB) such as Wikidata. EL is commonly formulated in three sequential steps: Named Entity Recognition (NER), where mentions are identified, Candidate Generation, where a list of possible entity candidates is generated, and Entity Disambiguation, where a final candidate is selected.

Earlier EL works relied on alias tables (dictionary from strings to possible Wikidata entities; often associated with a score) and key-word based retrieval methods \cite{spitkovsky2012cross,logeswaran2019zero, pershina2015personalized}. However, these approaches suffer on noisy text, such as short-form Tweets. An example of a difficult Tweet would be "\textit{\textbf{Liam} is a gr8 ML Researcher}" where the desired span to link would be "Liam". Here, an alias-based approach would only retrieve entities based on the span "Liam", of which there are 8,350 different Wikidata entities containing that name. Without the context of "gr8 ML Researcher", it quickly becomes unfeasible to find the correct candidate. Furthermore, alias based approaches are also heavily dependent on the spans retrieved, where the retrieved span must be exactly present in the alias table in order to be found \cite{spitkovsky2012cross, logeswaran2019zero, pershina2015personalized}. This presents a challenge due to the difficulties of NER systems on noisy social media text \cite{lample-etal-2016-neural, NERBias}. 

More recently, BERT-based dense entity retrieval approaches have shown to produce SOTA results on news datasets such as TACKBP-2010 and Mewsli-9 \cite{wu2019scalable,fitzgerald2021moleman,botha2020entity}. Dense retrieval approaches rely on relevant context around the mention, which is abundant in long and clean documents such as news, but often absent or brief in noisy and short user-generated text, such as that found on Twitter. 

Prior works that focus on social media linking, such as Tweeki \cite{harandizadeh2020tweeki}, used small, annotated datasets and did not study the more recent dense retrieval approaches. 

Recently, Twitter researchers released an end-to-end entity linking benchmark for Tweets called TweetNERD. It is the largest and most temporally diverse open-sourced dataset benchmark on Tweets \cite{mishra2022tweetnerd}. Excited by the availability of this benchmark, we study the application of recent linking methods on this large and noisy user generated data. We empirically evaluate sparse and dense retrieval approaches on this data and describe the challenges and design choices of building a robust linking system for Tweets.

Our main contributions are as follows:
\begin{enumerate*}[itemsep=0pt,parsep=0pt,label=\textbf{(\Alph*)}]
    \item To the best of our knowledge, we are the first study to compare dense retrieval, sparse retrieval, and lookup based approaches for Entity Linking in a social media setting, which makes our work relevant for the research community interested in processing noisy user generated text. 
    \item We assess the robustness of dense retrieval techniques in the presence of span detection errors coming from NER systems for social media text. This is a common problem for social media datasets as the top NER F1 score for social media datasets is significantly lower than other domains \cite{strauss2016results}.
    \item We assess the impact of using short Wikidata entity descriptions against the longer Wikipedia descriptions for representing candidates, and highlight the significant loss in performance from using shorter descriptions for social media text. This is relevant as many recent dense retrieval methods for generic Entity Linking have proposed using short descriptions from Wikidata for candidate representations. 
    \item Our analysis is the first to explore sparse and dense retrieval on the largest and most temporally diverse Entity Linking dataset for Tweets called TweetNERD \cite{mishra2022tweetnerd}.
    \item Finally, through quantitative and qualitative analysis, we assert the complimentary nature of candidates generated by lookup and dense retrieval based approaches. This asserts the validity of our hybrid approach towards candidate generation and is reflected in significant performance improvement by using hybrid candidate generation for Entity Linking.
\end{enumerate*}

\section{Methodology}

\subsection{Knowledge Base}
To represent our KB, we followed prior work and retrieved a July 2022 download of English Wikipedia\footnote{This was the latest version at the time of writing} \cite{wu2019scalable, de2020autoregressive}. 
However, Wikipedia also includes miscellaneous pages or pages that refer to multiple entities, such as disambiguation pages and "list of" pages. An example of such a page is "List of Birds of Canada" \footnote{https://en.wikipedia.org/wiki/List\_of\_birds\_of\_Canada}, which describes 696 distinct birds, each with their own respective Wikipedia page. 
To detect these pages, we retrieve the "instance of" category of each entity from Wikidata, which classifies each Wikipedia entity into distinct categories. Using this information, we reduce the entity set from 56.8M to 6.5M Wikidata entities.

\subsection{Span Detection}
\label{sec:span}
We observe the performance of our systems utilizing the Gold Spans provided by TweetNERD (Table \ref{tab:CGGold}) and compare that to using NER-based spans that reflect a more realistic use-case. The NER model is trained on Tweets from TweetNERD and is similar to the models described in \citet{lample-etal-2016-neural,NERBias}.

\subsection{Candidate Generation}

\subsubsection{Dense Retrieval}
Our dense retrieval approach retrieves candidates based on the similarity of tweet and entity embeddings. This is done by utilizing two separate language models to encode the semantic content of Tweets and Entities respectively. Our approach is motivated by \citet{wu2019scalable}, which utilized a similar strategy on a clean news corpus. Given a Tweet $t$ with mention span $s$ and entity $e^{i}$, we create dense embeddings as
\begin{align}
    T^{s} &= BERT_T([CLS]~t^{s}_l~[M_1]~span^{s}~[M_2]~t^{s}_r) \\
    E^{i} &= BERT_E([CLS]~title^{i}~[M_3]~desc^{i})
\end{align}
where $BERT_T$ and $BERT_E$ are two separate language models, $t^{s}_l$ and $t^{s}_r$ refer to the text to the left and right of the desired mention span $s$, and $title^{i}$ and $desc^{i}$ are the Wikipedia title and first ten sentences of the respective entity page. Finally, $[M_1]$, $[M_2]$, $[M_3]$ are special tokens to denote the separation of each of the fields in the input. 

Given these dense embeddings, we rank the pairing of entities $e$ to Tweet $t$ by computing the dot product between their corresponding CLS representations. During inference, we pre-compute the embeddings for every entity in our knowledge base and index them using fast $k$ nearest neighbour search provided by FAISS \cite{Johnson2017Billion}. We refer to this approach as Dense. 

\subsubsection{Sparse Retrieval}
\label{sec:lookup}
We utilize a traditional lookup-based approach for finding candidates as used by many prior works \cite{harandizadeh2020tweeki}. Specifically, we map surface forms to Wikipedia page candidates from the English Wikipedia parse of DBPedia Spotlight and rank candidates given $p(entity | surfaceForm)$. We also include Wikidata aliases and labels as both have been found previously to be beneficial for identifying named entities \cite{mishra-diesner-2016-semi,singh12:wiki-links,isem2011mendesetal} and entity candidates in text \cite{isem2011mendesetal,mishra2022tweetnerd,singh12:wiki-links}. We refer to this approach as Lookup. 

\section{Results}
\subsection{Experimental Setup}
We use TweetNERD for training and evaluation. It consists of 340K+ Tweets linked to entities in Wikidata \cite{mishra2022tweetnerd}. We follow the authors' setup and evaluate on TweetNERD-Academic and TweetNERD-OOD (out of domain), while the rest of the data is used for training. For Dense retrieval we use pre-trained BLINK\footnote{https://github.com/facebookresearch/BLINK} encoders which are trained on Wikipedia text and FAISS \cite{Johnson2017Billion} for indexing candidate embeddings. We compare that to a Lookup based system (Section \ref{sec:lookup}) and a BM25 baseline \cite{yang2018anserini}. For BM25, we utilize Wikipedia abstracts as candidate documents and mention spans as queries. 

In all experiments, we limit our retrieved candidates set for Dense and BM25 to the top 16 entities due to observed diminishing returns (Figure \ref{fig:recall-at-K}). For Lookup, we retrieve all exact match candidates since they are not explicitly ranked. As a result, the performance of Lookup reflects an upper-bound of the performance of that method. The average number of retrieved Lookup candidates is 19 while the median of 4, reflecting the long tail distribution of retrieved candidates per span. 

\subsection{Candidate Generation}
We begin by evaluating the impact of dense retrieval on Candidate Generation. Since we constrain our dense retrieval methods to 16 candidates, we measure Recall @16 of our various systems. 

\subsubsection{Gold Spans}
\begin{table}
    \centering
    \caption{Candidate Generation using Gold Spans (R@16)}
    \begin{tabular}{lrrrr}
    \toprule
        Data Split & Dense & Lookup & BM25 & Hybrid \\
        \midrule
        Academic & \underline{0.783} & 0.741 & 0.221 & \textbf{0.916} \\
        OOD & 0.772 & \underline{0.847} & 0.556 & \textbf{0.933} \\
        Overall & \underline{0.779} & 0.717 & 0.362 & \textbf{0.930} \\
        \bottomrule
    \end{tabular}

    \label{tab:CGGold}
\end{table}

We first observe the performance of our systems utilizing the Gold Spans provided by TweetNERD (Table \ref{tab:CGGold}). Contrasting Lookup and Dense, we can see that Dense outperforms on the Academic split by 4 points whereas Lookup outperforms on the Out-of-Domain split by 7.5 points. In addition, we see that our trivial BM25 baseline falls significantly behind with 0.221 recall on the Academic set and 0.556 on the OOD set.

\begin{figure}
    \centering
    \includegraphics[width=\linewidth]{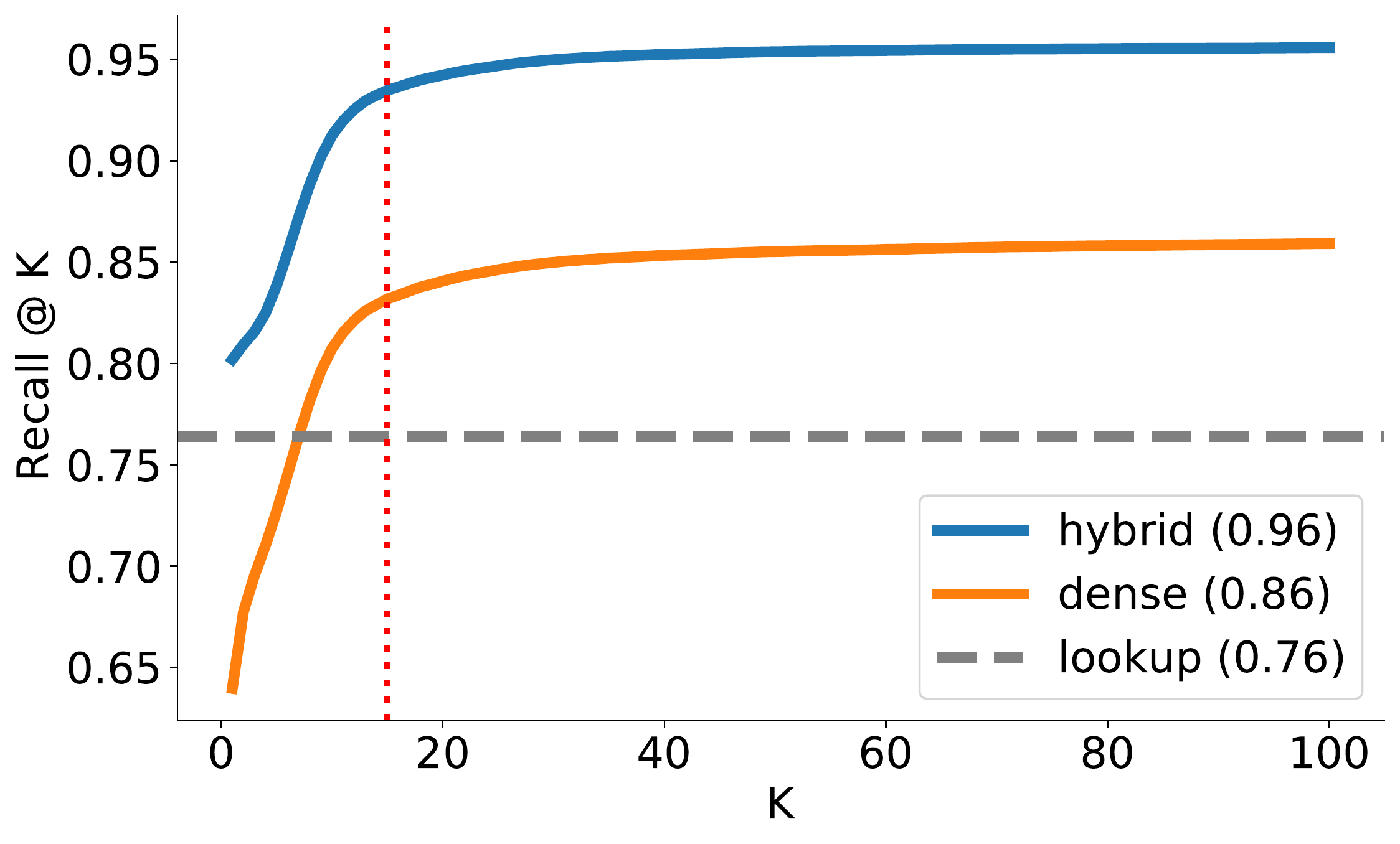}
    \caption{Recall @ K of Dense, Lookup and Hybrid using Gold Spans}
    \label{fig:recall-at-K}
\end{figure}

\begin{figure}
    \centering
    \includegraphics[width=\linewidth]{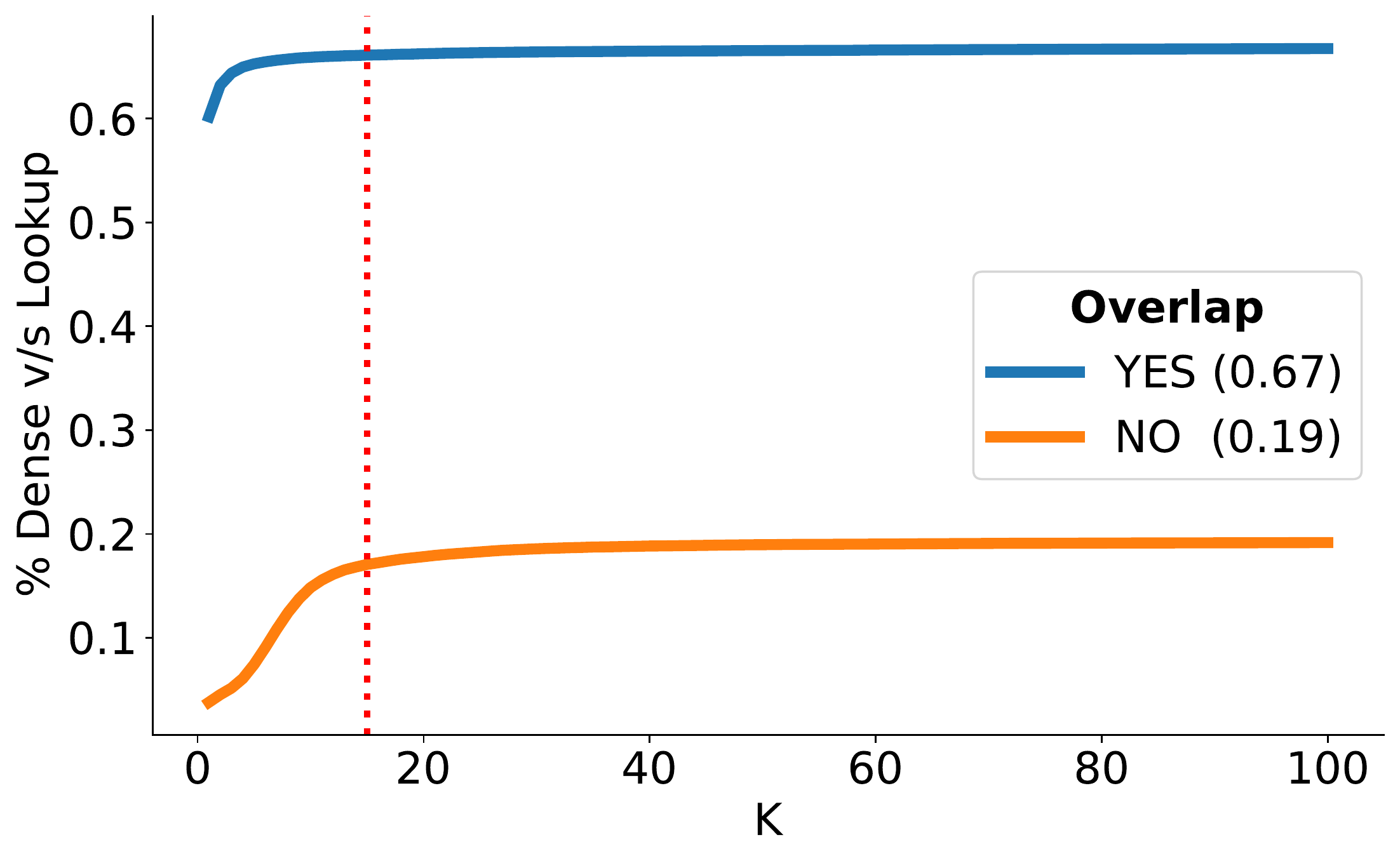}
    \caption{Overlap and Distinction of Dense v/s Lookup using Gold Spans}
    \label{fig:overlap-dense}
\end{figure}

\begin{table}
    \centering
    \caption{Unique Correct Candidates using Gold Spans}
    \begin{tabular}{lrrr}
    \toprule
        Data Split & Dense & Lookup & BM25 \\
        \midrule
        Academic & \textbf{7,719} & 5,268 & 1,043 \\
        OOD & 1,055 & \textbf{2,664} & 1,495 \\
        Overall & \textbf{8,774} & 7,932 & 2,538 \\
        \bottomrule
    \end{tabular}
    \label{tab:CGGold Unique}
\end{table}

\begin{table}
\centering
\caption{Candidate Overlap Across Lookup, Dense and BM25 using Gold Spans}
\begin{tabular}{cccrr}
\toprule
Lookup & Dense & BM25 & counts & prop \\
\midrule
     Y &     Y &    Y & 16,310 & 0.30 \\
     Y &     Y &    N & 19,810 & 0.36 \\
     Y &     N &    Y &  2,190 & 0.04 \\
     Y &     N &    N &  3,566 & 0.06 \\
     N &     Y &    Y &  1,079 & 0.02 \\
     N &     Y &    N &  8,298 & 0.15 \\
     N &     N &    Y &    361 & 0.01 \\
     N &     N &    N &  3,386 & 0.06 \\
\bottomrule
\end{tabular}

    \label{tab:cg-overlap-compare}
\end{table}

Upon further investigation, we find that Dense and Lookup methods produce mutually exclusive results. On the Academic dataset, we find that Dense retrieved 7719 unique correct candidates whereas lookup retrieved 5268 unique correct candidates (Table \ref{tab:CGGold Unique} and Table \ref{tab:cg-overlap-compare}). Leveraging these differences and inspired by \citet{van2020rel}, we take the union of both methods as a Hybrid approach. This approach yielded a significant \textbf{+17.5 recall} increase over Lookup and \textbf{+13.3 recall} increase over Dense on the Academic split. In Figure \ref{fig:recall-at-K}, we show the change in Recall for all approaches as K increases. We can see that the benefit of retrieving more Dense candidates plateaus after 16 candidates. However, we also find that candidates retrieved by Lookup and Dense continue to be mutually exclusive despite the larger candidate set (Figure \ref{fig:overlap-dense}). This illustrates that the performance plateau is not due to overlap in candidate sets but rather that both methods produce vastly different candidates. We investigate these differences in Section \ref{sec:qualitative}.

\subsubsection{NER Spans}
\begin{table}
    \centering
    \caption{Candidate Generation using NER Spans (R@16)}
    \begin{tabular}{lrrrr}
    \toprule
        Data Split & Dense & Lookup & BM25 & Hybrid \\
        \midrule
        Academic & \underline{0.761} & 0.613 & 0.164 & \textbf{0.880} \\
        OOD & 0.754 & \underline{0.757} & 0.440  & \textbf{0.903} \\
        Overall & \underline{0.759} & 0.715 & 0.245 & \textbf{0.887} \\
        \bottomrule
    \end{tabular}

    \label{tab:CGNER}
\end{table}
\begin{table}
    \centering
    \caption{Unique Correct Candidates using NER Spans}
    \begin{tabular}{lrrr}
    \toprule
        Data Split & Dense & Lookup & BM25 \\
        \midrule
        Academic & \textbf{8,362} & 4,711 & 983 \\
        OOD & 1,263 & \textbf{2,448} & 1,496 \\
        Overall & \textbf{9,625} & 7,159 & 2,479 \\
        \bottomrule
    \end{tabular}
    \label{tab:CGNER Unique}
\end{table}

Next, to reflect a real-life use-case, we investigate performance of our system on NER spans. Here, we annotate each Tweet using the NER service described in Section \ref{sec:span}. We capture the recall performance of our systems by evaluating the set of all retrieved candidates against the set of gold entities (Table \ref{tab:CGNER}). Here, we can see the benefits of Dense retrieval where Dense achieved similar performance on NER spans as utilizing gold spans. This is contrasted by Lookup, which realized a significant drop in performance. This is likely due to inaccuracies in our NER system, which can return spans that do not have exact entries in our pre-computed table. 

We also see a continuing trend of complementary results between Dense retrieval and Lookup. Here, Dense and Unique retrieved 8362 and 4711 unique correct entities on the Academic set, respectively (Table \ref{tab:CGNER Unique}). By combining the retrieved candidates from both sets, we can increase the performance of Lookup by $\approx$ \textbf{26.7 points} on all splits.

\subsubsection{Qualitative Analysis}
\label{sec:qualitative}
During our experiments, we found significant differences between the candidates retrieved by Dense retrieval and Lookup retrieval. We find that these differences can largely be categorized into span ambiguity, spelling, and the presence of context.

An example of a TweetNERD Tweet requiring context due to span ambiguity would be 
    \textit{"Wiz and \textbf{Amber}, Rihanna and Chris, Beyonce and jay-z \#grammyscouples"}
where the desired span is the word "Amber". 

In our results, we found that Lookup returned many entities containing the name "Amber", such as "AMBER Alert" (\qid{Q1202607}) and "Amber, Rajasthan, India" (\qid{Q8197166}), but not the correct entity "Amber Rose" (\qid{Q290856}). To the reader, it is clear upon reading the entire Tweet that the meaning does not concern a rescue service or city, but rather celebrities who have dated someone named "Wiz". This is contrasted by Dense retrieval, which returned the correct entity, but also similar entities such as celebrity "Amber Benson" (\qid{Q456862}). Furthermore, we can see in the Wikipedia entity description of Amber Rose that she had been married to Wiz Khalifa, information that would not be present in the lookup table. 

However, the presence of context can also be detrimental and misleading when taken literally. An example of such a TweetNERD Tweet would be 
    \textit{"No one here remembers The Marine and the \textbf{12 Rounds}."} 
where the desired span is "12 Rounds".

In this case, Dense retrieval returned incorrect candidates such as "12 Gauge Shotgun" (\qid{Q2933934}), instead of "12 Rounds" the movie (\qid{Q245187}). However, this was mitigated by Lookup, which accurately found the correct entity. We hypothesize that the context of "Marines" combined with "12 Rounds" misleads the Dense model to retrieve entities related to weaponry, instead of matching the literal title as Lookup did.

\section{Conclusion}
In this work, we have evaluated the usage of sparse and dense retrieval techniques towards candidate generation on social media text. In our qualitative and quantitative experimentation, we have highlighted the complementary strengths of both methods. Combined, our hybrid approach achieves significant improvements on TweetNERD, a large temporally diverse dataset for entity linking on Tweets. We also demonstrate the improvements that dense retrieval translates to improved downstream entity linking performance using both gold and NER based spans. 

There are also a few directions for future work. First, in this work we focused on the Candidate Generation step for Entity Linking. While we report preliminary results for the Entity Disambiguation step in Appendix Section \ref{sec:disambiguation}, future work could explore efficient ways to disambiguate the candidates retrieved from our hybrid approach. Second, future work could expand our evaluation beyond the English Tweets found in TweetNERD and develop a multi-lingual solution. Third, it is important to note that there are significant linguistic differences between the formal text found on Wikipedia and informal speech on Twitter. Recent work has explored leveraging mentions as entity descriptions, which could be applied to Twitter text to bridge this gap \cite{fitzgerald2021moleman}. 

Overall, our work highlights the best practices for improving entity linking on short and noisy social media text. We hope this work inspires future entity linking efforts on this challenging domain.
\bibliography{anthology,custom}
\bibliographystyle{acl_natbib}

\appendix
\section{Entity Disambiguation}
\label{sec:disambiguation}
To evaluate end-to-end EL performance, we conduct preliminary experiments by training a disambiguation model using the candidate set retrieved from our retrieval methods. Once we generate entity candidates, we score each <Mention, Entity> pair for each Tweet using common mention-entity Lookup based features (e.g., mention count per entity), entity only based features (e.g., Wikipedia page rank), and contextual mention-entity features generated by comparing the mention embedding in the text against the candidate entity description embedding. We train our model to identify the correct entity for each span among the retrieved candidates. Our architecture and features are like the ones described in \citet{kolitsas-etal-2018-end} with the major difference being the usage of a BERT based encoder instead of BiLSTM. 

While our focus is candidate generation, reporting end-to-end performance is important since improvement in candidate generation does not necessarily translate to end-to-end improvement. Dense, unlike Lookup, can retrieve the right candidate even when the mention span is missing due to NER errors, however our disambiguation system currently still requires a span in order to link a mention. 
\begin{table}
    \centering
    \begin{tabular}{lrr}
    \toprule
        Dataset Split & Dense & Lookup \\
        \midrule
        Academic & \textbf{0.617}     & 0.566  \\
        OOD      & \textbf{0.605}     & 0.568 \\
        Overall  & \textbf{0.610}     & 0.567 \\
        \bottomrule
    \end{tabular}
    \caption{F1 of Entity Disambiguation using NER Spans}
    \label{tab:NER Disambiguation}
\end{table}

Table \ref{tab:NER Disambiguation} shows the F1 score of our disambiguation model using candidates retrieved by our proposed methods. Our results demonstrate that the increased recall brought by Dense candidates have translated into increased end-to-end F1 on all splits when compared to Lookup, achieving a 0.04 F1 gain. Furthermore, we can see the largest difference on the Academic split, where Dense achieved 0.051 higher F1 then our lookup-based approach.

\section{Ablation Study}
\label{sec:ablation}
A core part of our methodology is how we represent entities. In our proposed approach, we utilize Wikipedia descriptions, which provide a verbose but rich description of entities. We refer to these descriptions as "Long" descriptions. To evaluate the impact of these descriptions on Dense and Lookup retrieval, we conduct an ablation study where we evaluate utilizing Wikidata descriptions. These descriptions are much shorter and terse, often never exceeding 5-6 words. An example of such a description would be "species of bird", which is shared by 23 828 different bird entities \footnote{https://www.wikidata.org/w/index.php?search=species+of+bird}. We refer to these Wikidata descriptions as "Short" descriptions. 

\begin{table}
    \centering
    \begin{tabular}{lrrr}
    \toprule
        Description & Recall & Precision & F1 \\
        \midrule
        \textbf{Lookup} & & & \\
        \midrule
         Short          & 0.484 & \textbf{0.686} & 0.567 \\
         Long      & 0.543 & 0.628 & 0.582 \\
        \midrule
        \textbf{Dense} & & & \\
        \midrule
         Short           & 0.299 & 0.249 & 0.272 \\
         Long       & \textbf{0.613} & 0.607 & \textbf{0.610} \\
        \bottomrule
    \end{tabular}
    \caption{Ablation Experiments on Entity Disambiguation}
    \label{tab:Ablation Study}
\end{table}

The results of our ablation study can be seen in Table \ref{tab:Ablation Study}. While we see an overall improvement when utilizing Long descriptions, the most significant impact can be seen on dense retrieval, where we see a leap of F1 performance from 0.272 to 0.610. Furthermore, we can also see that Lookup can still perform well when utilizing Short descriptions, achieving our highest precision result. 

There are a few reasons for these results. Due to the k-nearest neighbour nature of Dense retrieval, entities that are retrieved by this method are often very semantically similar. This was demonstrated in Section \ref{sec:qualitative}, where Dense retrieval returned a list of actors when trying to link to an actor mention. However, since short descriptions are often shared between related entities ("species of bird"), often the same description would appear in the retrieved list. This is contrasted by Lookup, where the list of retrieved entities is related only by mentioned name. As a result, the entities are typically much more diverse (AMBER Alert vs Amber Rose) and thus easier to disambiguate with shorter descriptions. 
\end{document}